\documentclass[conference]{IEEEtran}
\IEEEoverridecommandlockouts
\usepackage{cite}
\usepackage{amsmath,amssymb,amsfonts}
\usepackage{algorithmic}
\usepackage{graphicx}
\usepackage{textcomp}
\usepackage{xcolor}
\def\BibTeX{{\rm B\kern-.05em{\sc i\kern-.025em b}\kern-.08em
    T\kern-.1667em\lower.7ex\hbox{E}\kern-.125emX}}
\begin{document}

\title{Porimon: An LLM-Based Pokémon Battle Agent Enhanced by Long/Short-Term Knowledge Augmented Generation\\
}

\author{\IEEEauthorblockN{Dongyin Zhuo, Fengjunjie Pan, Nenad Petrovic, Alois Knoll}
\IEEEauthorblockA{\textit{Robotics, Artificial Intelligence and Real-Time Systems}\\
\textit{School of Computation, Information and Technology}\\\textit{Technical University of Munich}\\
Munich, Germany\\
\{dongyin.zhuo, f.pan, nenad.petrovic, k\}@tum.de
}
}
\maketitle

\begin{abstract}
In this paper, we use Pokémon Battles as a case study to investigate how to improve the performance of LLM-based agents in tasks that require opponent-aware planning without additional fine-tuning.
We propose Long/Short-Term Knowledge Augmented Generation (LSTKAG), a mechanism that enables Large Language Model-based (LLM-based) agents to leverage past states of the current task and retrieve experience summaries from similar previous task instances based on the current state.
Based on LSTKAG, we design Porimon, an LLM-based agent architecture for Pokémon Battles. To improve the decision making skill of the agent, we incorporate an external API for precise damage calculation and the more detailed game information retrieval.
We conduct tournament-like evaluation experiments comprising 15,000 battles for hyperparameter optimization, ablation studies, and performance evaluation. The results indicate that Porimon-based players with hyperparameter optimization significantly outperform players based on PokéLLMon, an LLM-based agent structure proposed in previous research, and the rule-based heuristic player.
Furthermore, our ablation study shows that Porimon variants outperform the one without extension in game information retrieval, which shows the contribution of that extension. However, the current experiment results are inconclusive regarding the contribution of Long-Term KAG.
These results suggest that introducing external resources, information from previous states of the current task, and experience summaries from similar previous task instances could elevate the performance of LLM-based agents designed for tasks requiring opponent-aware planning.

\end{abstract}


\section{Introduction}
Previous research has shown that Large Language Models (LLMs) can serve as zero-shot problem solvers~\cite{brown2020language, yao2022react}, showing their potential as an essential part of agents. Based on this finding, we attempt to improve the performance of agents by leveraging existing resources without any further training or fine-tuning. By leveraging external knowledge without additional training, our approach aims to enable the ability to incorporate information about unfamiliar tasks and environments into its decision-making process. Games are ideal testbeds for agents compared to the real world, because games are controllable and complex enough to capture certain aspects of real-world decision making while providing sufficient control over experimental conditions and avoiding the costs and risks of real-world experiments~\cite{hu2026surveylargelanguagemodelbased,yannakakis2018artificial}.

In our study, we choose Pokémon Battles as the test environment for agents to evaluate their ability to perform opponent-aware planning, because it challenges agents to make opponent-aware decisions regarding various strategies and limited actions. Existing approaches to this task could be divided into two categories. Agents based on Offline Reinforcement Learning (Offline RL), such as Metamon, require a lot of training data and are vulnerable to cases not included in the training data~\cite{grigsby2025humanlevelcompetitivepokemonscalable, riahi2026distribution}. In contrast, LLM-based agents can take advantage of their zero-shot problem-solving capabilities and externally retrieved knowledge to make decisions when facing unfamiliar tasks without additional training or fine-tuning. However, the agents based on LLMs suffer from hallucination and struggle with accurate damage calculation~\cite{huang2025survey,zhang2025mathematical,Fan2024RAGSurvey}. Retrieval Augmented Generation (RAG) enables LLMs to take advantage of external related information during inference~\cite{Fan2024RAGSurvey}. PokéLLMon further deployed Knowledge Augmented Generation (KAG) for game information retrieval, and In-Context Reinforcement Learning (ICRL) to incorporate information from previous turns within the current battle into planning~\cite{hu2024pokellmonhumanparityagentpokemon}. However, neither of these existing implementations explicitly stores and retrieves reusable experience summaries from completed battles to support decision-making in future battles.

To address these limitations, we propose Porimon, an LLM-based Pokémon battle agent built upon PokéLLMon~\cite{hu2024pokellmonhumanparityagentpokemon}. Porimon enhances the agent structure with more detailed game information retrieval, including precise damage calculation through the Smogon Damage Calculation API~\cite{smogon_damage_calc}, and introduces a Long/Short-Term Knowledge Augmented Generation (LSTKAG) mechanism that lets the agent reason over previous turns within a battle and accumulate reusable experience across battles. We evaluate Porimon in a multistage tournament against PokéLLMon-based and rule-based heuristic players.

Our main contributions are as follows:

\begin{enumerate}
    \item We propose Porimon, an LLM-based Pokémon battle agent that integrates extended game information retrieval, including API-based damage calculation, with the LSTKAG mechanism. This enables the agent to exploit information from previous turns in a battle and to learn from finished battles without extra training or fine-tuning, while mitigating hallucination and calculation errors.
    \item We conduct multistage tournament experiments with 15,000 battles in total, covering hyperparameter optimization, ablation studies, and comparison tests, and evaluate the players using the Bradley--Terry model and Wald tests with Holm--Bonferroni correction~\cite{bradley1952rank,wald1943tests,holm1979simple}.
    \item Experimental results indicate that Porimon-based players statistically significantly outperform both PokéLLMon-based and heuristic players. The ablation study shows that the extension in game information retrieval contributes to this advantage while the contribution from Long-Term KAG remains inconclusive. These findings suggest that providing LLM-based agents with precise information matters.
\end{enumerate}
\section{Methodology}

\begin{figure*}[tb]
    \centering
    \includegraphics[width=0.8\textwidth]{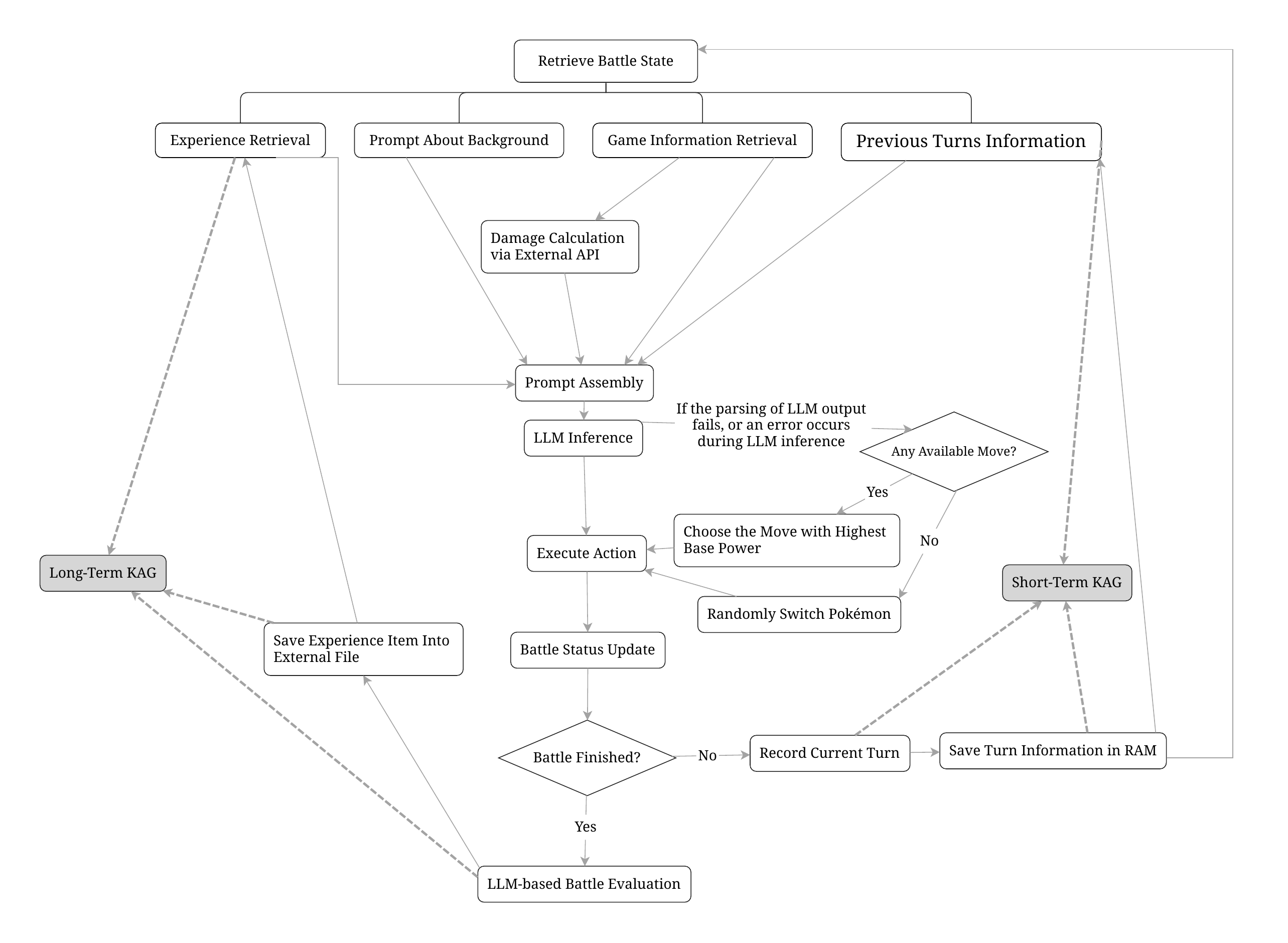}
    \caption{The workflow of a Porimon-based agent during and after a single Pokémon Battle game. In this figure, the solid lines describe the working process of the player, and dashed lines illustrate the components that belong to the Long-Term KAG or Short-Term KAG mechanism.}
    \label{fig:agentplayerworkflow}
\end{figure*}

Figure~\ref{fig:agentplayerworkflow} illustrates the workflow of a Porimon-based agent during and after a battle.

\subsection{Game Information Retrieval}
Built based on PokéLLMon, our agent first retrieves the game information about all known Pokémon from Smogon, an open-source Pokémon database built for Pokémon Showdown, the framework we used for the evaluation experiment, upon receiving the current battle state~\cite{pokemon_showdown}. The contents we retrieve include but are not limited to:
\begin{itemize}
\item Information about all known Pokémon in current state.
\item Description of moves our Pokémon could use.
\item The effectiveness relationship between types.
\end{itemize}

In addition to the implementation of PokéLLMon, we also retrieve the information about abilities and carried items of each Pokémon for making more informed decisions.

Moreover, Smogon Damage Calculation API is used to calculate accurate damage ranges for all available moves of our Pokémon, in order to help the agent identify edge cases and account for randomness.

\subsection{Short-Term Knowledge Augmented Generation}
\label{stkag}
As developed in PokéLLMon, during the battle, after each event, our agent transforms the original message from the Pokémon Showdown server to natural language texts for inclusion in the prompt. These text segments are generated according to a fixed policy and concatenated together. These concatenated texts from different turns are then separated by the delimiter [sep]. We use a sliding window with size $T$ to include information from up to $T$ previous turns.

\subsection{Long-Term Knowledge Augmented Generation}
\label{ltkag}
After each battle, our agent summarizes the battle using the recording created in the process of Section~\ref{stkag} focusing on the following fields:
\begin{itemize}
\item Short summary of the battle.
\item Tactical lessons learned from the battle.
\item An importance score of each Pokémon in the battle.
\item The exact information of Pokémon in both teams.
\end{itemize}

For example, an experience item could record a battle where a \textit{Rattata} carrying a \textit{Focus Sash} used \textit{Endeavor} and \textit{Quick Attack} to beat our last Pokémon (summary). Then our agent could identify this strategy as a tactical lesson and give this Rattata a high importance score. Moreover, the detailed information about the Pokémon in both teams is saved.

For the retrieval of the summaries, we represent each Pokémon as a vector containing base stats information and remaining HP, and a set including its type, ability, and status condition. Using this representation, we can calculate the similarity of two Pokémon by the following equation:

\begin{equation}
  w(p_1,p_2) = \alpha e^{-||v_1-v_2||_2} + (1-\alpha)\frac{|S_1\cap S_2|}{|S_1\cup S_2|},
\end{equation}

where $v$ and $S$ represent the vector and set representations of a Pokémon respectively, i.e., we denote a Pokémon as $p_i=(v_i,S_i)$. The Euclidean-distance term compares numerical attributes, and the Jaccard similarity term compares categorical attributes~\cite{jaccard1901etude}. $\alpha \in (0,1)$ is a hyperparameter balancing their contributions. 

Considering that the order of Pokémon in a team does not have an impact on the game process, we use the following equation to calculate the similarity between teams:

\begin{equation}
\begin{aligned}
  S(A,B) &=\frac12(D(A,B)+D(B,A))\\
         &=\frac1{2|A|}\sum_{a\in A}\max_{b\in B} w(a,b)+\frac1{2|B|}\sum_{b\in B}\max_{a\in A} w(b,a)
\end{aligned}
\end{equation}

$A\text{ and }B$ denote two teams, which could be represented as a set of Pokémon. Although $w$ is symmetric, $D(A,B)=D(B,A)$ does not always hold. To obtain a symmetric team similarity measure, we average $D(A,B)$ and $D(B,A)$.

Based on the team similarity, we calculate a similarity score between the current battle state and each experience item in the memory. Let $C_{\text{our}}$ and $C_{\text{opp}}$ denote our team and the opponent's team in the current state, and $M_{\text{our}}$ and $M_{\text{opp}}$ denote the corresponding teams recorded in the experience item. We implemented two approaches:
\begin{enumerate}
    \item Uniform approach: 
    \begin{equation}
        s=(1-\lambda)\,S(C_{\text{our}},M_{\text{our}})+\lambda\, S(C_{\text{opp}},M_{\text{opp}}),
    \end{equation}
    where $\lambda$ controls the relative weight of the opponent's team and is manually set to $0.7$.
    
    \item Weighted approach: In order to account for the fact that different Pokémon may contribute unequally to the battle, we introduce an importance-weighted similarity measure. We denote the rating of each Pokémon from memory as $r_p$. Then we use the following equation to calculate the similarity between teams regarding the importance of Pokémon:
    \begin{equation}
    S'(A,B,M_1,M_2)=\frac{R(A,M_1)+R(B,M_2)}{N},
    \end{equation}
    where $R(A,M)=\sum_{m\in M} r_m\max_{o\in A}w(m,o)$ is the importance-weighted similarity of the memorized team $M$ to the current team $A$, and $N=\sum_{m\in M_1}r_m+\sum_{m\in M_2}r_m$ normalizes the score by the total importance. We then have the weighted similarity:
    \begin{equation}
    \begin{aligned}
    s' = \frac12\Bigl(&\frac12\bigl(D(C_{\text{our}},M_{\text{our}})+D(C_{\text{opp}},M_{\text{opp}})\bigr) \\
    &+S'(C_{\text{our}},C_{\text{opp}},M_{\text{our}},M_{\text{opp}})\Bigr)
    \end{aligned}
    \end{equation}
    Since importance ratings are available only in stored experience items, the weighted calculation can only be performed from the perspective of the memory. To keep the score approximately symmetric, we average it with the unweighted similarity from current state side.
\end{enumerate}

Battle summaries are generated from the last 12,000 characters of the log of each battle. Once 10 experience items are stored, the three highest-scoring items are retrieved, and their summaries and tactical lessons are included in the prompt.

\subsection{Exception Handling}

Although we request a JSON-formatted response, the LLM may generate a response that cannot be parsed correctly or choose an invalid action. Our agent may make up to $T_r$ generation attempts, where $T_r$ is manually set. Once the LLM produces a valid response, the action is executed accordingly. If all the generated responses are invalid, our agent chooses the move with the highest base power. If our agent has to switch Pokémon (e.g., our current Pokémon is fainted), it selects the replacement at random. 
\section{Experiment Design}
\begin{figure}[t]
    \centering
    \includegraphics[width=\linewidth]{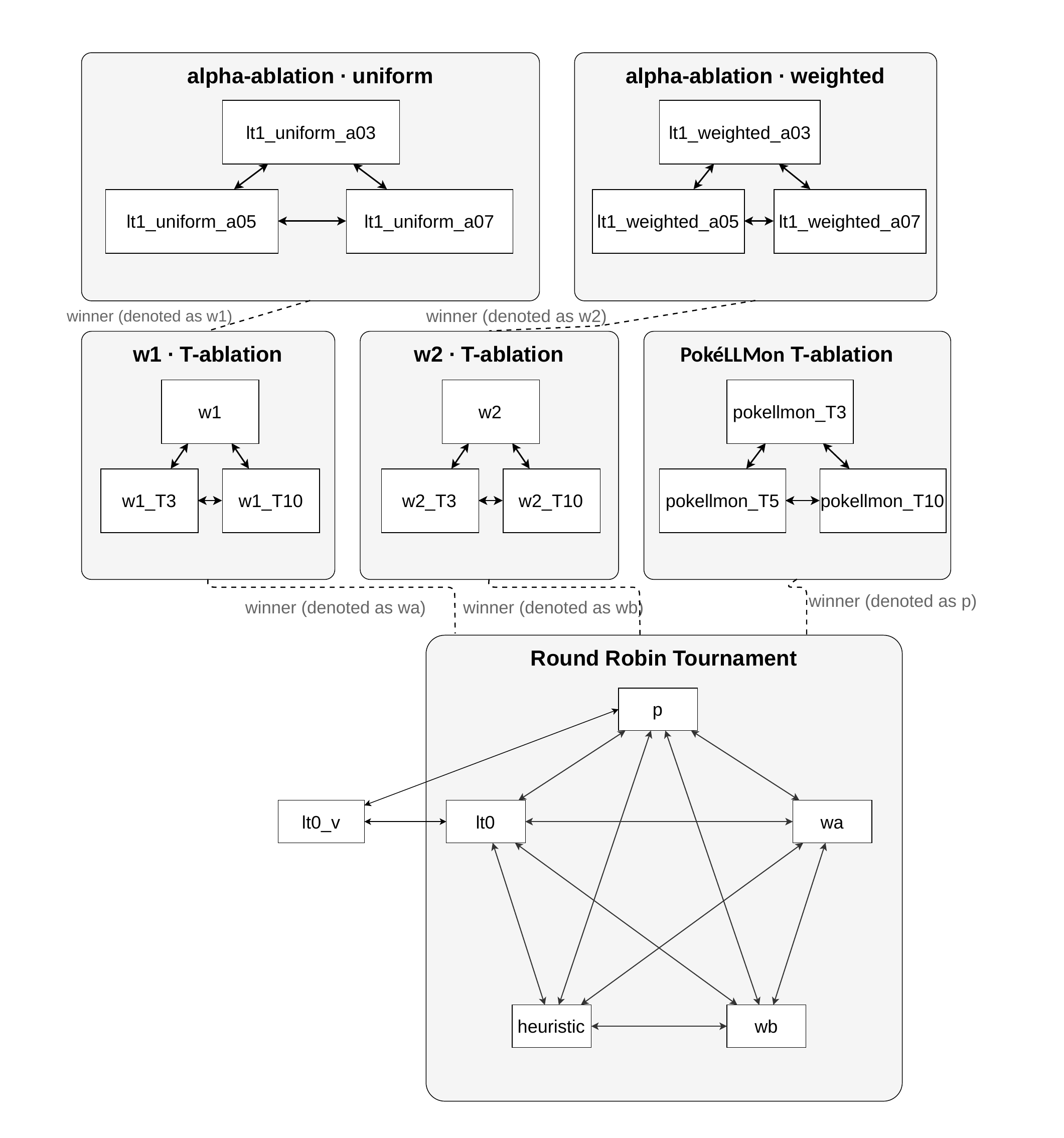}
    \caption{Graphical description of the tournament style experiment for performance evaluation.}
    \label{fig:tournament}
\end{figure}
\subsection{Environment Configuration}
We performed all our battles on a locally deployed Pokémon Showdown server~\cite{pokemon_showdown}. We selected Gen8 Random Battle as the game environment to reduce the influence of team composition on the experimental results, since all the Pokémon are randomly assigned to both players. 

In addition, Dynamax is a mechanism introduced in Gen 8 that temporarily increases both the maximum and current HP of a Pokémon and grants access to Max Moves for three turns. We disabled it during our evaluation to simplify the environment and remain consistent with PokéLLMon~\cite{hu2024pokellmonhumanparityagentpokemon}.
\subsection{Player Configuration}
Due to limited budget, we set $T_r=1$ to reduce the number of LLM requests. 

We used the GPT-5.6 Luna model with default sampling parameters as the LLM component of all
the LLM-based agent players. Due to limited budget, we set the reasoning effort to none.

For the precise damage calculation, we also deployed the Smogon Damage Calculation API locally~\cite{smogon_damage_calc}.

The player configurations are encoded in their names with the following conventions:
\begin{itemize}
    \item The default hyperparameter setting of $T$ is 5.
    \item The prefixes lt1, lt0, pokellmon, and heuristic denote Porimon-based players with and without Long-Term KAG feature, PokéLLMon-based players, and rule-based heuristic player, respectively.
    \item For lt0 players, the suffix \_v means the extension to game information retrieval is disabled, and the suffix \_w means only the extension part of game information retrieval is enabled, i.e., the game information which PokéLLMon-based players also retrieve is removed.
    \item For lt1 players, the suffixes \_uniform and \_weighted indicate the memory retrieval strategy used for Long-Term KAG as mentioned in Section~\ref{ltkag}.
    \item The suffixes \_T$n$ and \_a$0m$ mean we set the hyperparameter as $T=n$ and $\alpha=\frac{m}{10}$.
    \item Players named p, w1, w2, wa, and wb are the winners of the hyperparameter optimization stages; each inherits the configuration of the winning variant when they do not have any suffix. If a suffix is present, it overrides the corresponding parameter in the inherited configuration.
\end{itemize}

\subsection{Multistage Tournament Evaluation}
As described in Figure~\ref{fig:tournament}, our evaluation is conducted in three stages:
\begin{enumerate}
    \item The optimization of hyperparameter $\alpha$ for Porimon-based players, resulting in winners w1 and w2.
    \item The optimization of hyperparameter $T$ for all LLM-based players, resulting in winners wa, wb, and p.
    \item A round-robin tournament among heuristic, p, lt0, wa, and wb for the comparison test, where lt0\_v additionally plays against p and lt0 for the ablation study on the basic prompt and Long-Term KAG.
\end{enumerate}

To make the ablation study on game information retrieval more accurate, we also performed an additional round-robin tournament among lt0, lt0\_v, and lt0\_w.

For all the match-ups referred to above, 500 battles are performed, resulting in 30 match-ups and 15,000 battles in total. In each match-up, the Porimon-based players with Long-Term KAG feature started with an empty, match-up-specific memory, preventing information leakage across match-ups.

\subsection{Statistical Methods}
For hyperparameter selection, we select the variant with the highest average win rate in each round-robin tournament. Since pairwise outcomes are not necessarily transitive, selecting a winner solely through pairwise comparisons may be unreliable. 

We use the Bradley-Terry model to evaluate the performance of each player in the multistage tournament experiment, using the heuristic player as the reference player~\cite{bradley1952rank}. Using the MLE and covariance matrix calculated from that process, we perform pairwise and linear-contrast Wald tests to compare the performance of players~\cite{bradley1955rank,wald1943tests}. To control the error rate, we used the Holm-Bonferroni method to correct the significance test result~\cite{holm1979simple}. 

For the additional round-robin tournament, we use a binomial Test to evaluate whether a player statistically significantly outperforms another. The Holm-Bonferroni method is also used here for correction. We also calculate the 95\% confidence interval by the Wilson score interval, because it shows high robustness in edge cases~\cite{wilson1927probable,brown2001interval}.
\section{Experiment Results}
\subsection{Hyperparameter Optimization}
For the optimization of $\alpha$, no evaluated agent significantly outperforms another. For the uniform variant of Porimon-based players, the best-performing $\alpha$ among our tested values is 0.7, and for the weighted variant the value is 0.3.

For both variants of Porimon-based players as well as PokéLLMon-based players, the optimal $T$ is 10. For PokéLLMon-based players, the player with $T=10$ statistically significantly outperforms the one with $T=5$ ($p_{adj}\approx0.025$). All the other match-ups show no statistically significant difference.
\subsection{Comparison Test}
\begin{figure*}[tb]
    \centering
    \includegraphics[width=\textwidth]{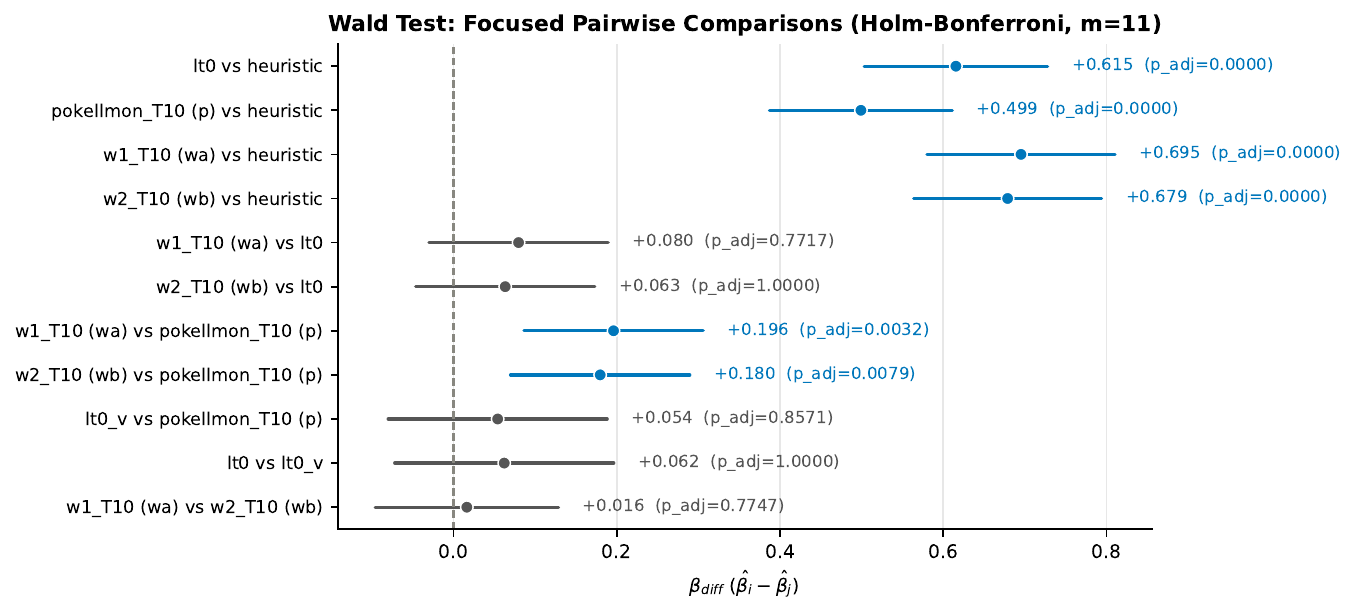}
    \caption{The significance test focusing on the performances of players in each match-up. The dot shows the performance difference of the two players in a match-up, and the bar illustrates the 95\% CI. The  statistically significant performance differences are marked in blue.}
    \label{fig:wald}
\end{figure*}

Figure~\ref{fig:wald} illustrates the results of the Wald tests. This figure indicates the following findings:
\begin{enumerate}
    \item All agents based on both Porimon and PokéLLMon architectures statistically significantly outperform the heuristic player after hyperparameter optimization ($p_{adj}<5\times10^{-5}, \beta_{diff}\geq 0.499$).
    \item Both uniform and weighted variants of Porimon-based players significantly outperform PokéLLMon-based players ($p_{adj}\approx0.003, \beta_{diff}\approx0.196$ for uniform variant; $p_{adj}\approx0.008, \beta_{diff}\approx0.180$ for weighted variant).
    \item The two variants of Porimon-based players show no statistically significant difference in their performance ($p_{adj}\approx0.775, \beta_{diff}\approx0.016$).
\end{enumerate}
\begin{figure}[bt]
    \centering
    \includegraphics[width=\linewidth]{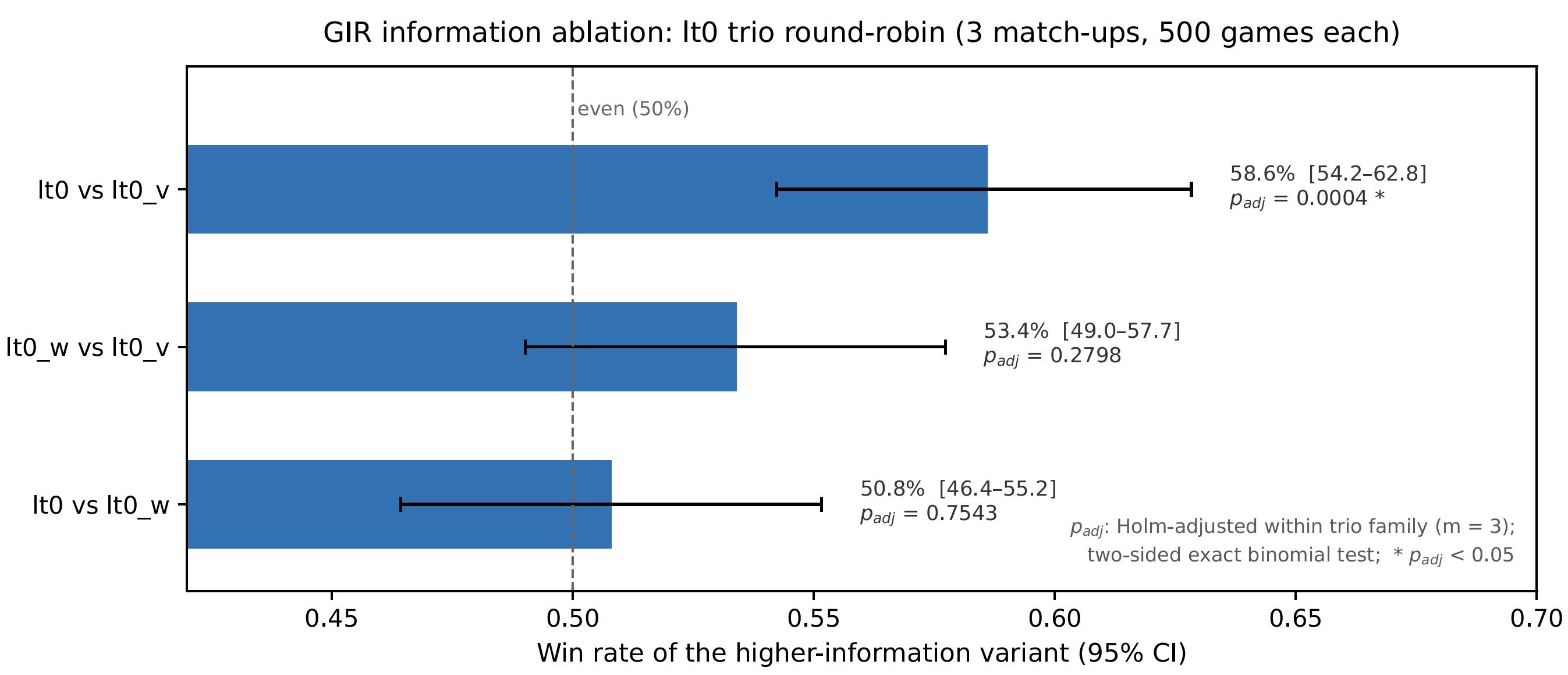}
    \caption{The win rate of each lt0 variant in the round-robin tournament. The blue bars show the win rate of each player, and the black bars show the 95\% CI.}
    \label{fig:trio}
\end{figure}
\subsection{Ablation Study}
Figure~\ref{fig:wald} supports the following observations:
\begin{enumerate}
    \item Extended game-information retrieval increases the performance of the Porimon-based agent, but the difference is not statistically significant in the multistage tournament ($p_{adj}\approx1.000,\beta_{diff}\approx 0.062$).
    \item The Porimon variants with Long-Term KAG do not significantly outperform the one without Long-Term KAG feature ($p_{adj}\approx0.772, \beta_{diff}\approx0.080$ for uniform variant; $p_{adj}\approx1.000, \beta_{diff}\approx0.063$ for weighted variant).
\end{enumerate}

In the additional round-robin tournament among the three lt0 variants, we observe that lt0 significantly outperforms lt0\_v ($p_{adj}\approx 4\times 10^{-4}$) while the performances between lt0 and lt0\_w and the performance difference between lt0\_v and lt0\_w are not statistically significant ($p_{adj}\approx 0.754$ for lt0 and lt0\_w; $p_{adj}\approx 0.280$ for lt0\_v and lt0\_w), confirming the contribution of game information retrieval extension. The difference in statistical significance between the multistage tournament and the round-robin tournament may be due to randomness or the difference in the number of hypotheses included in the Holm–Bonferroni correction. The result of this tournament is shown in Figure~\ref{fig:trio}.

\begin{figure}[t]
    \centering
    \includegraphics[width=\linewidth]{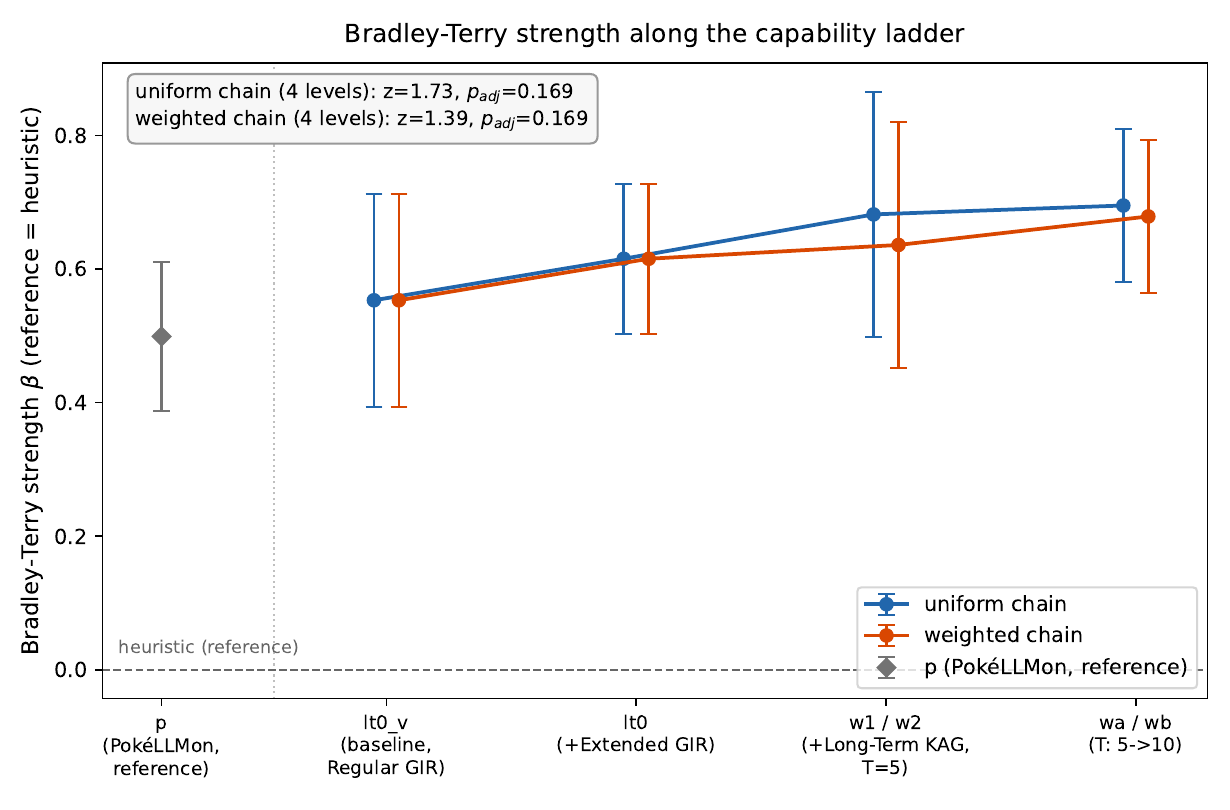}
    \caption{Bradley--Terry strength along the component ladder, tested by Wald tests of linear contrasts with Holm--Bonferroni correction ($m=2$). Blue and orange lines show the derivation of the uniform and weighted Porimon variants from the PokéLLMon player; vertical bars show 95\% CIs relative to the heuristic player.}
    \label{fig:chain}
\end{figure}

In addition, Figure~\ref{fig:chain} shows that when a new component is added to the agent or a hyperparameter optimization is done, the performance of our Porimon implementation increases monotonically across the evaluated configurations. Nevertheless, this effect is not statistically significant ($p_{adj}\approx0.169$ for both Porimon variants).
\section{Discussion and Conclusion}
In this paper, we present Porimon, an LLM-based agent architecture for Pokémon Battles enhanced by LSTKAG, designed to improve its performance by exploiting external resources and the experience gained from previous tasks without extra training or fine-tuning.

Our experimental results show that, after hyperparameter optimization, Porimon-based agents statistically significantly outperform those based on PokéLLMon, our reference baseline, as well as the rule-based heuristic player. The additional round-robin experiment indicates that the extension of game information retrieval contributes significantly. However, even though the performance of Porimon players improves with the addition of the Long-Term KAG feature, the observed improvements were small and not statistically significant, leading to the fact that the current experimental results remain inconclusive regarding its contribution. 

Even though Porimon demonstrates considerable performance, we acknowledge several limitations of our implementation and experiment:
\begin{enumerate}
    \item The number of battles per match-up in the experiment may be insufficient.
    \item For the Long-Term KAG, all the experience items are retained indefinitely. In the future, it is possible to employ cache-replacement policies such as least frequently used (LFU) to accelerate the retrieval process.
    \item For the implementation of Long-Term KAG, the retrieval is solely dependent on the similarity between the experience item and the current state, regardless of whether the information in the item is useful or not.
\end{enumerate}

Future research could address these limitations and evaluate the Porimon architecture in other games that require opponent-aware planning. Such an evaluation would require adapting game-specific knowledge retrieval and state-similarity metrics while retaining the LSTKAG framework.

\section*{Acknowledgment}
We sincerely thank the Smogon community for providing the Smogon Damage Calculator API and the Pokémon Showdown platform, which made the implementation and evaluation in this paper practical and accurate.

\bibliographystyle{IEEEtran}
\bibliography{bibliography}

\vspace{12pt}
\end{document}